%% file: main.tex
\documentclass[letterpaper]{article} % DO NOT CHANGE THIS
\usepackage[submission]{aaai23}  % DO NOT CHANGE THIS
\usepackage{times}  % DO NOT CHANGE THIS
\usepackage{helvet}  % DO NOT CHANGE THIS
\usepackage{courier}  % DO NOT CHANGE THIS
\usepackage[hyphens]{url}  % DO NOT CHANGE THIS
\usepackage{graphicx} % DO NOT CHANGE THIS
\usepackage{natbib}  % DO NOT CHANGE THIS AND DO NOT ADD ANY OPTIONS TO IT
\usepackage{caption} % DO NOT CHANGE THIS AND DO NOT ADD ANY OPTIONS TO IT
\usepackage{newfloat}
\usepackage{listings}
\DeclareCaptionStyle{ruled}{labelfont=normalfont,labelsep=colon,strut=off} % DO NOT CHANGE THIS
\input{preamble}
\input{sec/00_meta}

\begin{document}

\maketitle

\input{sec/01_abtract}

\input{sec/1_introduction}

\input{sec/2_related}
\input{sec/3_analysis}

\input{sec/4_method}

\input{sec/5_experiment}

\input{sec/6_conclusion}

%%%%%%%%% REFERENCES
{
    % \clearpage
    % \small
    % \bibliographystyle{aaai23}
    \bibliography{aaai23}
}

\end{document}

%% file: preamble.tex
\usepackage{overpic}
\usepackage{enumitem} %< control spacing in itemize/enumerate/...
\usepackage{overpic} %< add raw math symbols to figures
\usepackage{color}
\usepackage{amsmath}
\usepackage{booktabs}
\usepackage{multirow}
\usepackage{multicol}
\usepackage{amssymb}
\usepackage{floatrow}
\newfloatcommand{capbtabbox}{table}[][\FBwidth]
\definecolor{turquoise}{cmyk}{0.65,0,0.1,0.3}
\definecolor{purple}{rgb}{0.65,0,0.65}
\definecolor{dark_green}{rgb}{0, 0.5, 0}
\definecolor{orange}{rgb}{0.8, 0.6, 0.2}
\definecolor{red}{rgb}{0.8, 0.2, 0.2}
\definecolor{darkred}{rgb}{0.6, 0.1, 0.05}
\definecolor{blueish}{rgb}{0.0, 0.3, .6}
\definecolor{light_gray}{rgb}{0.7, 0.7, .7}
\definecolor{pink}{rgb}{1, 0, 1}
\definecolor{greyblue}{rgb}{0.25, 0.25, 1}

\usepackage{blindtext}

\renewcommand{\paragraph}[1]{\vspace{1em}\noindent\textbf{#1}.}
\newcommand{\proposed}{our method}
\newcommand{\Metric}{\textit{mean-angular-deviation}}
\newcommand{\metric}{$mav$}
\newcommand{\figref}[1]{Fig.~\ref{#1}}
\newcommand{\tabref}[1]{Table~\ref{#1}}
\newcommand{\eqnref}[1]{Eq.~\eqref{#1}}
\newcommand{\secref}[1]{Sec.~\ref{#1}}

\newcommand{\etal}{{\emph{et al.}}}
\newcommand{\ie}{{\emph{i.e.}}}
\newcommand{\eg}{{\emph{e.g.}}}

\usepackage{bbm}
\usepackage[ruled]{algorithm}
\usepackage{algpseudocode}

%% file: sec/00_meta.tex
\title{Jitter Does Matter: Adapting Gaze Estimation to New Domains}
\author{
    Ruicong Liu\textsuperscript{\rm 1}~
    Yiwei Bao\textsuperscript{\rm 1}~
    Mingjie Xu\textsuperscript{\rm 1}~
    Haofei Wang\textsuperscript{\rm 2}~
    Yunfei Liu\textsuperscript{\rm 1}~
    Feng Lu\textsuperscript{\rm 1,2,}\thanks{Corresponding Author.}\\
    {\textsuperscript{\rm 1} State Key Laboratory of VR Technology and Systems, 
		School of CSE, Beihang University}  \\
	{\textsuperscript{\rm 2} Peng Cheng Laboratory, Shenzhen, China} \\
	\small{\texttt{\{liuruicong, baoyiwei, xumingjie, lyunfei, lufeng\}@buaa.edu.cn}}~
	\small{\texttt{ wanghf@pcl.ac.cn}}
}
\affiliations{
    \textsuperscript{\rm 1}Association for the Advancement of Artificial Intelligence\\
    1900 Embarcadero Road, Suite 101\\
    Palo Alto, California 94303-3310 USA\\
    publications23@aaai.org
}

%% file: sec/01_abtract.tex
\begin{abstract}
% Deep neural networks have demonstrated superior performance on appearance-based gaze estimation tasks. However, existing studies focus less on cross-domain gaze estimation. Due to variations in person, illuminations, and background, gaze estimation performance usually degrades dramatically. To tackle such issues, recent studies have pointed out that using adversarial attack will help improve cross-domain adaptability. In this paper, we innovatively combine adversarial attack with contrastive learning for domain adaptation. Specifically, we propose the Contrastive learning with Adversarial samples for Unsupervised Gaze Adaptation (\proposed). First, we generate adversarial samples by introducing adversarial attack. Then, a contrastive learning framework is employed to keep the consistency between the raw and adversarial samples. Finally, we propose a validation-based post-processing method to reduce the deviation on the target domain. Experiments demonstrate that the \proposed~ outperforms the state-of-the-art approaches on four gaze domain adaptation tasks. Our work offers new insight for cross-domain gaze estimation tasks, and has the potential to be used in real-world gaze estimation applications.

Deep neural networks have demonstrated superior performance on appearance-based gaze estimation tasks. However, due to variations in person, illuminations, and background, performance degrades dramatically when applying the model to a new domain. In this paper, we discover an interesting gaze jitter phenomenon in cross-domain gaze estimation, \ie, the gaze predictions of two similar images can be severely deviated in target domain. This is closely related to cross-domain gaze estimation tasks, but surprisingly, it has not been noticed yet previously. Therefore, we innovatively propose to utilize the gaze jitter to analyze and optimize the gaze domain adaptation task. We find that the high-frequency component (HFC) is an important factor that leads to jitter. Based on this discovery, we add high-frequency components to input images using the adversarial attack and employ contrastive learning to encourage the model to obtain similar representations between original and perturbed data, which reduces the impacts of HFC. We evaluate the proposed method on four cross-domain gaze estimation tasks, and experimental results demonstrate that it significantly reduces the gaze jitter and improves the gaze estimation performance in target domains.

\end{abstract}

%% file: sec/1_introduction.tex
\section{Introduction}
\label{sec:intro}

Gaze indicates the direction along which a person is looking. It has been adopted in various applications, such as semi-autonomous driving\cite{A:demiris2007prediction, A:majaranta2014eye, A:park2013predicting} and human-robot interaction\cite{A:admoni2017social, A:terziouglu2020designing, A:wang2015hybrid}. With an increasing demand for predicting user intent implicitly, appearance-based gaze estimation has attracted more attention recently. To train the gaze estimator using deep learning neural networks, a number of large-scale datasets have been proposed~\cite{G:zhang2020eth, G:zhang2017mpii, G:funes2014eyediap, G:kellnhofer2019gaze360}.

However, due to variations in subjects, backgrounds, and illuminations, the performance of deep learning-based gaze estimation algorithms deteriorate significantly when applying the model trained in one dataset to new datasets. Recently, several techniques have been applied to address this cross-domain problem, such as adversarial learning\cite{D:tzeng2017adda, D:cui2020gvbgd}, few-shot learning\cite{G:park2019few, D:yu2019improving} and self-training\cite{D:cai2020generalizing}.
Among them, unsupervised domain adaptation (UDA) method\cite{D:wang2019gazeadv, G:kellnhofer2019gaze360, D:liu2021generalizing} is one of the promising approaches that attracts much attention. While requiring no labels makes it more applicable to real-world scenarios, it also makes the task more challenging.

% The goal of UDA in gaze estimation is to adopt a model to new domains,

% 修改版
Existing approaches usually optimize the gaze accuracy during adaptation directly. Instead, we design an approach that starts with the analysis of a phenomena we observed that occurs in crossing domain tests. Where we can look for the factors that cause the problems, and the factors can then be used as guidance for us to find a more explainable solution for domain adaptation.
%原始版
%Existing approaches usually optimize the gaze accuracy during adaptation directly, instead, here we begin to think about an approach that starts with the analysis of the phenomena that occur when crossing domains. Where we can look for the factors that cause the problems, and the factors can then be used as guidance for us to find a more explainable solution for domain adaptation.
% byw版
%Previous studies usually optimize the gaze accuracy during adaptation directly. In this paper, we study cross domain gaze estimation problem from a phenomenon we observed. We find the reason behind this phenomenon through several prove-of-concept experiments and design our method based on the conclusion of analysis.

% In order to make the model better applied to the target domain, our study starts from the observed problem: jitter. In our observation, we find that jitter is a significant problem, especially when crossing domains, which means two very similar images can be predicted with gazes far apart (shown in \figref{fig:jitter}). This can make the gaze estimation model unstable in the real world, thus reducing the practicality. Therefore, it is meaningful to study the cross-domain problem from the analysis of jitter.

% \input{fig/teaser}

In this paper, we observe the gaze jitter phenomena: two very similar images could be predicted with gazes severely deviated (shown in \figref{fig:jitter}), particularly when crossing domains. As shown in \figref{fig:jitter}, on the test set in the source domain, the model gives similar predictions when the input images are similar. In contrary, in the target domain, even if the input images are very similar, the model may still give predictions that are severely deviated. In this paper, we name this phenomenon as gaze jitter, and in addition, we consider gaze jitter as a manifestation of gaze error across domains, and use this phenomenon as a starting point to find a solution for domain adaptation.

Based on the above observation, we start to analyze why the gaze jitter phenomenon occurs and discover an important factor, i.e., the high-frequency component (HFC), which introduces gaze jitter problem and lowers the gaze estimation accuracy. Inspired by this, we propose our gaze adaptation framework. At first, our framework adds additive HFC to the input data, then it employs contrastive learning to keep the consistency between the original data and the perturbed data, thus making the model learn features with less impact of high-frequency component.
Our method leads to significant jitter reduction and performance improvement on various cross-domain gaze estimation tasks. The primary contributions of this paper are summarized as follows:

\begin{itemize}[leftmargin=*]
\setlength\itemsep{-.3em}
\item For the first time, we discover the gaze jitter problem on cross-domain gaze estimation tasks. We find that high-frequency component is an important factor introducing jitters. 
\item We propose a framework for cross-domain gaze estimation that suppresses the influence of high-frequency component, resulting in less jitter and better cross-domain gaze estimation accuracy.
\item Experimental results demonstrate that our method exhibits exceptional performances on four gaze domain adaptation tasks using only a small number of target images.
\end{itemize}

\input{fig/jitter}

%% file: fig/jitter.tex
\begin{figure}[t]
\begin{center}
% \begin{overpic} 
% [width=\linewidth]
% {example-image-a}
% \end{overpic}
%\includegraphics[width=\linewidth]{fig/jitter3.pdf}
% \includegraphics[width=\linewidth]{fig/teaser.pdf}
\includegraphics[width=\linewidth]{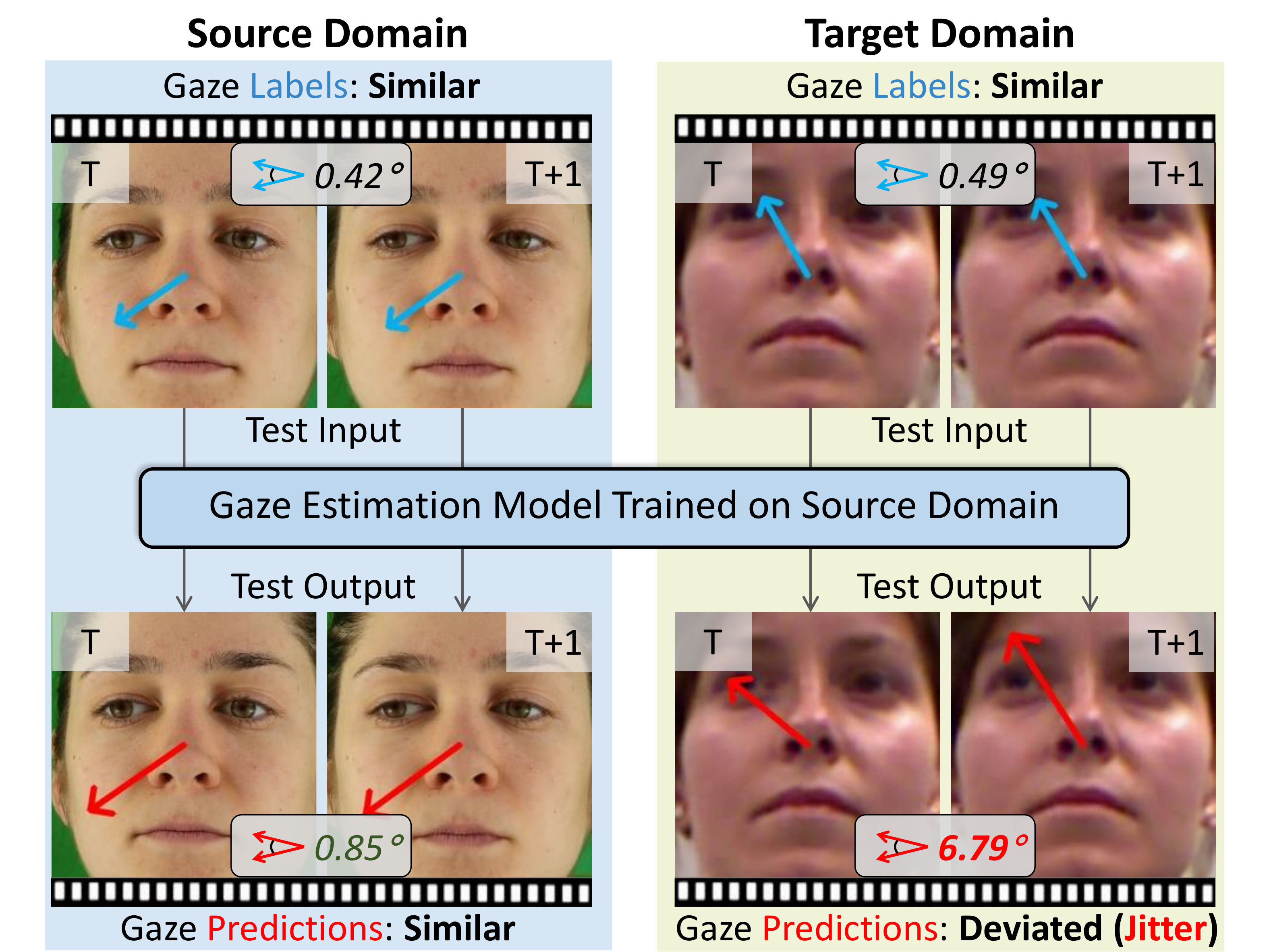}
\end{center}
\caption{
% 
%Illustration of gaze jitter in target domains (MPIIGaze, EyeDiap): two very similar images could be predicted with gazes far apart. The gaze predictions are predicted by the model trained in source domains.
We observe \textbf{gaze jitter} during cross-domain gaze estimation. Even though similar input images are expected to output close gaze directions, the predicted output can be severely deviated in target domain (bottom-right). We find such a gaze jitter a good indicator to help analyze and optimize cross domain gaze estimation.
}
\label{fig:jitter}
\end{figure}

%% file: sec/2_related.tex
\section{Related Work}

\subsection{Appearance-based gaze estimation}
Appearance-based gaze estimation aims to predict the human gaze from appearance. Zhang \etal~ proposed the first CNN-based gaze estimation method \cite{G:zhang2017mpii}, which uses eye images. With the release of many large-scale gaze datasets\cite{G:zhang2020eth, G:funes2014eyediap, G:kellnhofer2019gaze360, G:zhang2017mpii}, appearance-based gaze estimation has attracted more and more attention. Many methods have been proposed to estimate accurate gazes on public datasets.\cite{G:cheng2020coarse, D:guo2020dagen, D:shrivastava2017learning, D:wang2019gazeadv}.

However, most studies focus on the gaze estimation within a single dataset\cite{G:lu2014adaptive, G:park2019few, D:yu2019improving}. Due to the diversity of different datasets, almost all gaze estimation methods suffer from poor cross-domain capability\cite{G:cheng2020coarse, D:wang2019gazeadv}. Recent works \cite{D:liu2021generalizing, G:zhang2020eth} investigated the cross-domain capability of a gaze estimator, which improves the applicability to real-world scenes. 
% In this paper, we also propose an unsupervised domain adaptation framework for generalizing to new domains.

\subsection{Unsupervised domain adaption}
Unsupervised domain adaption(UDA) is a transfer learning task that requires no target labels. Previous UDA approaches can be divided into three categories: discrepancy, reconstruction, and adversarial methods. Discrepancy methods aim to minimize the domain gap using some distance metrics, such as Maximum Mean Discrepancy (MMD) \cite{D:ghifary2014domain}
% , Joint Maximum Mean Discrepancy (JMMD) \cite{D:long2017deep}, Central Moment Discrepancy (CMD) \cite{D:zellinger2017central},
and Local Maximum Mean Discrepancy (LMMD) \cite{D:zhu2020deep}. Reconstruction methods \cite{D:glorot2011domain, D:bousmalis2016domain} use a reconstruction strategy that allows a model to learn features from both domains\cite{D:wang2018deep}. Adversarial methods are inspired by the generative adversarial network (GAN)\cite{D:goodfellow2014generative}. In \cite{D:ganin2015unsupervised, D:tzeng2017adda, D:cui2020gvbgd, D:yu2019transfer}, they make a domain discriminator and a generator play a min-max game, thereby explicitly reducing the distance between the source and target domains. 
% In this paper, the domain gap is minimized in an adversarial manner.

However, most existing UDA methods have been designed for classification or semantic segmentation tasks. Gaze estimation is a typical regression task, its continuous label space makes it even more challenging.

\subsection{Adversarial attack}
The goal of the adversarial attack is to generate adversarial noise. Although this noise is a type of high-frequency component that usually cannot affect human cognition, recent studies\cite{N:goodfellow2014fgsm, N:szegedy2013intriguing} have shown that deep neural networks are highly vulnerable to it.
Although some adversarial attack methods have been proposed \cite{N:moosavi2016deepfool, N:su2019one} in the past few years, they mainly follow two ideas proposed by \cite{N:goodfellow2014fgsm} and \cite{N:madry2017pgd}.
Recently, the adversarial attack has been applied to UDA tasks in various fields\cite{A:ma2021understanding, D:yang2021gta5, D:madry2017towards, A:liu2021training}, which reminds us of the potential of applying it to the field of gaze estimation.

% In this paper, we employ adversarial attacks to generate adversarial samples, which are used to form positive pairs in contrastive learning.

\subsection{Contrastive learning}
 On UDA tasks, contrastive learning is usually used to help the model learn better representations. It encourages augmentations of the same input to have more similar representations compared to augmentations of different inputs. 
Common contrastive learning frameworks achieve this aim by constructing two kinds of pairs: positive pairs containing similar instances and negative pairs containing different instances. Then it maximizes the consistency over the positive pairs and pushes apart samples from the negative pairs.
Recent contrastive learning studies, \eg, Memory Bank\cite{C:wu2018unsupervised}, MoCo\cite{C:he2020moco}, SimCLR\cite{C:chen2020simclr}, and PCL\cite{C:li2020pcl} have reached considerable improvement on some downstream tasks.

Contrastive learning has been used for UDA in some tasks, such as action recognition \cite{A:kang2020contrastive} and semantic segmentation \cite{A:liu2021domain}. The significant effect shows the capability of contrastive learning to learn useful representations.

%% file: sec/3_analysis.tex
\section{Motivation}

% As observed in \figref{fig:jitter}, jittser is a significant problem in gaze estimation, which means two very similar images could be predicted with gazes far apart. 
% In this section, we analyze the additive jitter brought by cross-domain and its factor. Then, we propose the idea to develop a framework for improving cross-domain performance which is based on the factor of jitter and able to reduce it. By the way, the new metric \metric~ to measure the magnitude of jitter is defined in this section.
% In this section, we first introduce a new metric \metric, which is designed to measure the magnitude of jitter in gaze estimation. Then, we use it to compare the jitter magnitude on the source and target domains and find that the jitter problem is much more significant in the target domain. After that, we study the impact of HFC and find it an important factor of jitter. We also give an analysis of how the HFC causes jitter. Following the analysis, the jitter-reducing method is introduced.
In this section, we observe that gaze jitter is a significant phenomenon in cross-domain gaze estimation. Then, we discover one important factor introducing jitter: high-frequency component. Finally, we design a framework to adapt the gaze estimation to new domains with the guidance of this discovery.

% \subsection{Observation of Gaze Jitter}
\subsection{Gaze Jitter: an Observation in Cross-Domain Gaze Estimation}
As we know, the performance of pre-trained gaze estimation model usually degrades in unseen target domains. Most previous works study this problem only focusing the rise of gaze estimation error in target domains.
However, along with the rise in error, we observe the gaze jitter phenomenon, \ie, images with similar appearances and labels can be predicted with severely deviated gaze directions. 
% The gaze jitter is neglected in previous cross-domain gaze estimation works, which usually only focus on the improvement gaze accuracy.
% To find intuitive phenomena of performance degradation, we pre-train models on the source domains and directly test them in target domains. Then, we analyze the performance in target domains and here give our observation. 
% As shown in \figref{fig:jitter}, we observe from the testing results that gaze jitter occurs: two very similar images can be predicted with gazes far apart, especially when crossing domains. 
In our further analysis, we find the jitter is significantly larger in the target domain than in the source domain (\figref{fig:jitter}). To numerically measure the magnitude of jitter, we define a new metric \Metric~ (\metric) as follow.
\begin{itemize}
\item[]
\textit{Definition.} 
We denote the test dataset as $\mathcal{D}=\{(x_i,y_i)|_{i=1}^{N_{\mathcal{D}}}\}$, where $x_i$
and $y_i$ denote the i-th image and the corresponding gaze direction, ${N_{\mathcal{D}}}$ is the number of images. $\{\hat{y}_i|_{i=1}^{N_{\mathcal{D}}}\}$ is the prediction from an estimation model. 
We design \metric~ to measure the magnitude of jitter:
% , so it should reflect the difference in predictions between similar inputs. Toward this end, we define \metric~ as:

    \begin{equation} \label{eq:local-consistency}
    \begin{aligned}
    mav&\,(\mathcal{D}) = \frac{1}{N} \sum_{x_i, x_j \in \mathcal{D}} |\langle \hat{y}_i,\hat{y}_j \rangle - \langle y_i,y_j \rangle|,\\
    &s.t. \; SSIM(x_i, x_j)>\alpha,\, \langle y_i,y_j \rangle<\beta.
    \end{aligned}
    \end{equation}

\noindent where $\langle y_i,y_j \rangle$ indicates the angle between $y_i$ and $y_j$, $N$ is the number of image pairs that satisfy the constraint, and $SSIM$ measures the similarity between images \cite{A:wang2004ssim}. 
The \metric~ calculates the deviation between the angles of the predictions and labels from image pairs with similar appearance and labels.
% The constraint is used to restrict the calculation of \metric~ to images that are similar to each other, since gaze jitter means similar input images are predicted with gazes far apart. 
Empirically, we set $\alpha=0.75$ and $\beta=1^\circ$.

% As gaze jitter means two very similar input images are predicted with gazes far apart, defining "similar" is a prerequisite. Here we define "similar" as:
    % \begin{equation} \label{eq:similar}
    % \begin{aligned}
    % SSIM(x_1, x_2)>0.75;\; \overset{\wedge}{y_1,y_2}<1^{\circ},
    % \end{aligned}
    % \end{equation}

% Since the \metric~ is designed to measure the magnitude of jitter, it should reflect the difference in predictions between similar inputs. Toward this end, we define \metric~ as:

% \noindent where $\hat{y}_1, \hat{y}_2$ are the predictions of $x_1, x_2$ from the gaze estimation model. Note that the \metric~ is only calculated between previously defined "similar" image pairs. 

\end{itemize}

To briefly verify the correctness of \metric, we add random noise (Gaussian) with gradually increasing level to the test data, the change of \metric~ is illustrated in \figref{fig:local-consistency}(a). The \metric~ reflects the trend of magnitude correctly, \ie, stronger noise results in stronger jitter.

Using the \metric, we measure the magnitude of jitter on both source and target domains. Results are shown in \figref{fig:local-consistency}(b), 
% the value sees a significant increment when applying the models to the target domains, which means
and the gaze jitter is indeed more significant on the target domain. Therefore, we treat gaze jitter as a good indicator to help analyze and optimize cross-domain gaze estimation.

% Particularly, the ETH-XGaze (E) \cite{G:zhang2020eth} and Gaze360 (G) \cite{G:kellnhofer2019gaze360} dataset are employed as source domains, and the MPIIGaze (M) \cite{G:zhang2017mpii} and EyeDiap (D) \cite{G:funes2014eyediap} dataset are employed as target domains since E and G have a much larger data distribution than M and D.
% We directly apply models pre-trained on the source domains for testing, and the \metric~ results are shown in \tabref{tab:jitter}. The value sees a significant increment when applying the models to the target domains, which also shows that the jitter is indeed more significant in the target domain.
% \input{tab/jitter}

% \paragraph{Measurement of jitter in gaze estimation}
% Here we define \Metric~ (\metric). The \metric~ is designed to measure the magnitude of jitter, so it should reflect the difference in predictions between some very similar inputs. At first, we have to define the criteria of what is "similar". Assume that we have two images and their gaze labels $(x_1, y_1), (x_2, y_2)$. Here we define "similar" as:

% Then, after getting the predictions of $x_1, x_2$ and denote them as $\hat{y}_1, \hat{y}_2$, the \metric~ is defined as:
\input{fig/local-consistency}
\input{fig/low-freq}

% \subsection{Does HFC Introduce Gaze Jitter?}
\subsection{Why Does Gaze Jitter Occur?}
\label{sec:high-freq}
The question that arises is: \textit{Why does gaze jitter occur?} 
% According to \cite{N:wang2020high}, there are some components in an image that are imperceptible to a human affects the generalizability of CNNs, a typical example is HFC.
According to \cite{N:wang2020high}, CNNs may capture high-frequency component (HFC) that are misaligned with human visual preference. This conclusion is coincide with the observed gaze jitter: images which are similar to human (similar appearance) could be very different to CNN (severely deviated gaze predictions). Therefore, we speculate HFC could also be a factor introducing jitter.

To verify this idea, we filter out information of target domain images from high to low-frequency by Fourier transform during testing and see the influence. 
Particularly, the proportion of information filtered out gradually increases (from 0 to 100\%). As shown in \figref{fig:low-freq}, filtering out HFC reduces both the gaze jitter and error, which proves the conjecture that HFC is one of the factors introducing gaze jitter. In addition, the results in \figref{fig:local-consistency}(a) also support this conjecture, since HFC is also added when adding random noise. Based on this discovery, reducing the impact of HFC can be an effective direction to improve cross-domain accuracy and reduce gaze jitter.

\subsection{Domain Adaptation by Contrastive Learning} \label{sec:design}
To reduce the impact of HFC, we propose to utilize contrastive learning. 
In contrastive learning, positive pairs are generated from a given sample by data augmentations. 
Then, contrastive loss pulls the features of positive pairs closer.
As a result, the model extracts better feature and learns better generalization ability \cite{A:kang2020contrastive}. The model extracts similar feature from positive pairs and learns to neglect irrelevant differences between positive pairs caused by data augmentation.
% As a result, the model extracts similar feature from positive pairs. The model learns to neglect the impact of data augmentation thus learns better generalization ability \cite{A:kang2020contrastive}.

Accordingly, we propose to generate such positive pairs from target domain images by adding HFC.
Under the constrain of contrastive loss, gaze estimation model extracts similar features for positive pairs \ie~original image and image with additive HFC. 
In this way, the impact of HFC are reduced. Consequently, gaze jitter in target domain is reduced and the generalization ability to target domain is improved. 

The final question is what type of HFC should be used. In this paper, we show that using adversarial noise to generate positive pairs is effective.
First, adversarial noise is a form of HFC \cite{N:zhou2021high, N:olivier2021high}. Second, previous work \cite{N:wang2020high} proves that adversarial vulneralbility is a indicator when CNN captures HFC. Our experiments show that adversarial noise outperforms other data augmentation methods in \secref{sec:ablation-study}.

%--------------------------------------------------------------------------------------

%% file: fig/local-consistency.tex
\begin{figure}[t]
\begin{center}
% \begin{overpic} 
% [width=\linewidth]
% {example-image-a}
% \end{overpic}
\includegraphics[width=\linewidth]{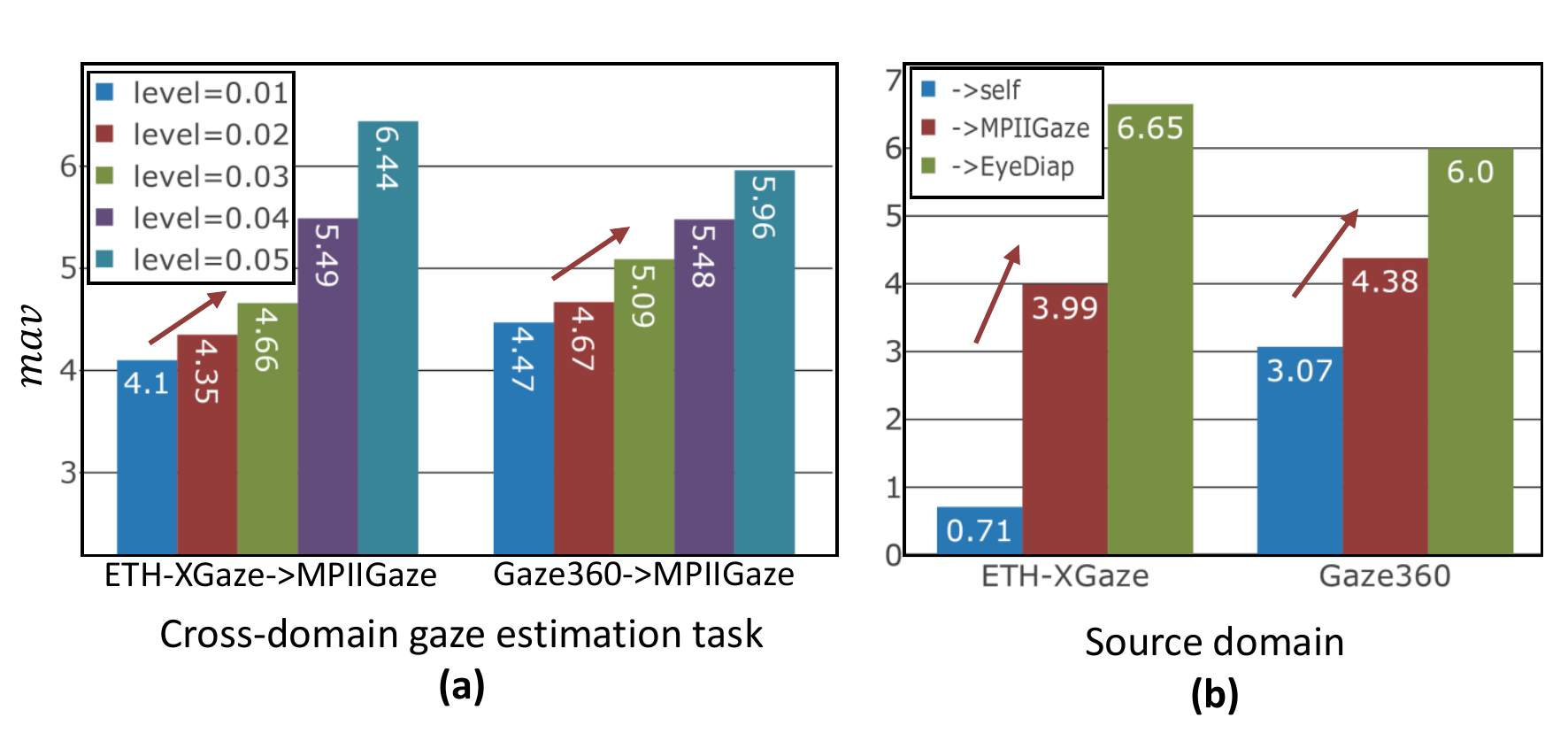}
\end{center}
\caption{
(a) The magnitude of jitter measured by \metric~ when adding random noise (Gaussian) with gradually increasing scale to the test data. (b) The magnitude of jitter on the source and target domains. The ETH-XGaze and Gaze360 datasets are employed as source domains, and the MPIIGaze and EyeDiap datasets are used as target domains.
}
\label{fig:local-consistency}
\end{figure}

%% file: fig/low-freq.tex
\begin{figure}
\begin{center}
\includegraphics[width=\linewidth]{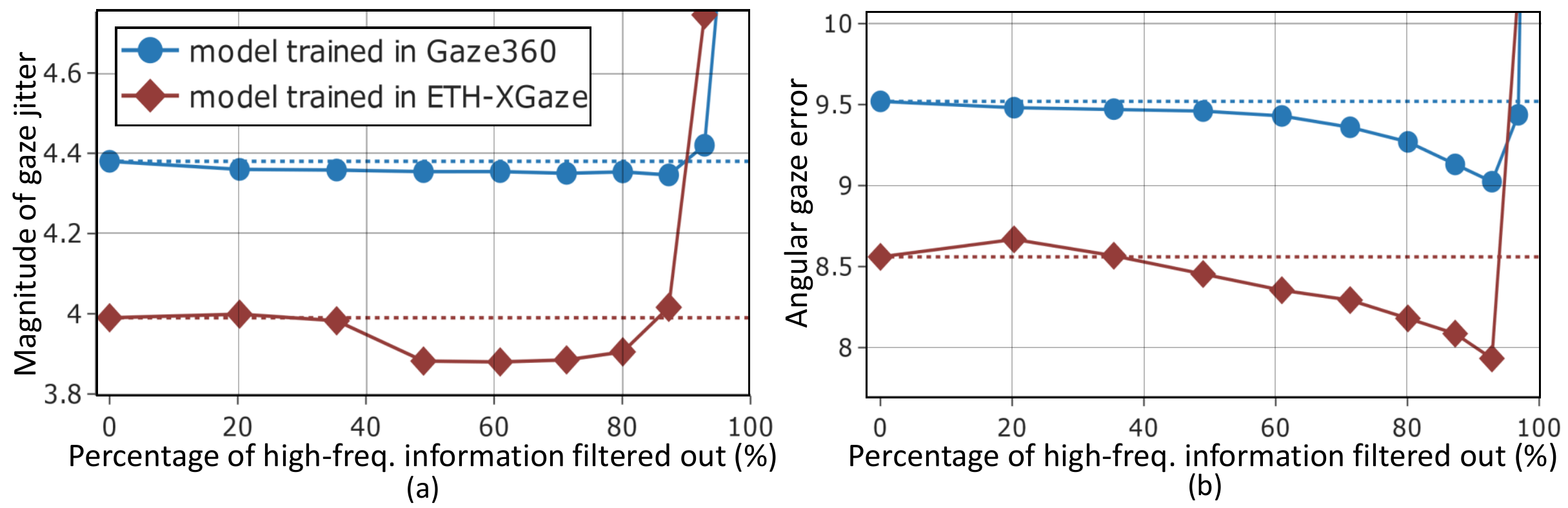}
\end{center}
\caption{
(a) and (b) illustrate the level of gaze jitter and angular gaze error on the target domain (MPIIGaze) using low-pass images, respectively. X axis indicates the percentage of high-frequency information filtered out, and the dotted lines illustrate the results when giving original images as input.
}
\label{fig:low-freq}
\end{figure}

%% file: sec/4_method.tex
\input{fig/overview}

\section{Method}

\subsection{Overview}
First in the data augmentation phase (\secref{sec:attack}), we use adversarial attacks to add high frequency information to the data and generate positive pairs. Then, the contrastive learning module will reduce the impact of HFC by optimizing the contrastive loss (\secref{sec:con}). Finally, we follow the idea of commonly-used adversarial learning to adapt the model to the target domain (\secref{sec:adv}).
An overview of the architecture is described in \figref{fig:overview}.

\subsection{Data Augmentation for Contrastive Learning}\label{sec:attack}
    \subsubsection{Adversarial Attack}\label{sec:preliminary}
%     Recall that the objective of adversarial attacks is:
    
%     \begin{equation} \label{eq:attack}
% 	\mathop{\arg\max}\limits_{\eta}\mathcal{L}(f(x+\eta), y),
%     \end{equation}
    
    % \noindent where $f$ abstracts the model into a function, $\eta$ is allowed perturbations, $x' \gets x+\eta$ is an adversarial sample of $x$ with adversarial noise $\eta$.
    According to \secref{sec:design}, adversarial attack is a good choice for data augmentation. Therefore, we use adversarial attacks to add adversarial noise and generate positive pairs.
    Existing adversarial attacks mainly follow two different ideas, the first one is fast gradient sign method (FGSM)\cite{N:goodfellow2014fgsm}:
    
    \begin{equation} \label{eq:fgsm}
	x' = x + \epsilon \cdot sign(\bigtriangledown_x\mathcal{L}(f(x), y)),
    \end{equation}
    
    \noindent where $\epsilon$ is the magnitude of the perturbation. This attack is a simple one-step scheme. 
    By contrast, another idea is to use the multi-step variant, which is essentially projected gradient descent (PGD) \cite{N:madry2017pgd}:
    
    \begin{equation} \label{eq:pgd}
	x^{t+1} = \mathbf{\Pi}_{x+S}(x^t + \epsilon \cdot sign(\bigtriangledown_x\mathcal{L}(f(x), y))),
    \end{equation}
    
    \noindent where $\mathbf{\Pi}$ is the projection function.
    In this paper, we simultaneously use FGSM and PGD to generate adversarial noise. 
    % which is a type of HFC. 
    Although the noise usually cannot affect human cognition, it can easily fool deep neural networks.

    \subsubsection{Adding High-Frequency Component}\label{sec:attack}
    % In this section, we use the adversarial attack as described in \secref{sec:preliminary} to add adversarial noise as a type of HFC to the input images. 
    Here we use adversarial attack described before for data augmentation, where HFC is added to the data.
    As shown in \figref{fig:overview}, \proposed~ takes source-target pairs $(x^s, x^t)$ as inputs. The gaze estimation network $G(\cdot|\theta_G)$ contains a feature extractor $F(\cdot |\theta_G)$, which follows a multi-layer perceptron.
    % $\Pi(\cdot |\theta_G)$, thus the network $G$ can be expressed as $F(\cdot |\theta_G) \circ \Pi(\cdot |\theta_G)$.
    Note that the parameters $\theta_G$ loaded by $G$ before adaptation are pre-trained with the source domain data $\mathcal{D}_s$.
    
    Before adaptation, with the target domain data $\mathcal{D}_t$ as input, the network $G(\cdot |\theta_G)$ would generate pseudo labels $\{{y}^t_i |_{i=1}^{N_t}\}$ by simply forward propagation. At each training iteration, a batch of $B$ source-target pairs are randomly sampled from $\mathcal{D}_s$ and $\mathcal{D}_t$, resulting in pairs: $\{(x_i^s, x_i^t) |_{i=1}^B\}$.  Their augmented samples $\{({x'}_i^s, {x'}_i^t) |_{i=1}^B\}$ are generated as follow:
    
    \begin{equation} \label{eq:img-aug}
    {x'}_i^{s,t}=\left\{
    \begin{aligned}
        FGSM(x_i^{s,t}, y_i^{s,t}, \mathcal{L}_{gaze}) & , & 0.5, \\
        PGD(x_i^{s,t}, y_i^{s,t}, \mathcal{L}_{gaze}) & , & 0.5.
    \end{aligned}
    \right.
    \end{equation}
    
    \noindent where $FGSM$ and $PGD$ are defined as \eqnref{eq:fgsm} and \eqnref{eq:pgd}. During adaptation, each image is augmented with 50-50 probability using one of both methods. $\mathcal{L}_{gaze}$ is usually defined as L1 loss:
    
    \begin{equation} \label{eq:gazeloss}
    \mathcal{L}_{gaze}(x,y;\theta_G)=||G(x |\theta_G)-y||_1.
    \end{equation}
    
    % $\mathcal{L}_{dis}$ in \eqnref{eq:augment} is the loss to encourage a binary discriminator $D(\cdot |\theta_D)$ to distinguish data from the source and target domains, which is defined as:
    
    % \begin{equation} \label{eq:disloss}
    % \begin{aligned}
    % \mathcal{L}_{dis}= &-log(1-D(F(x^s) |\theta_D))-logD(F(x^t) |\theta_D)\\
    % &-log(1-D(F({x'}^s) |\theta_D))-logD(F({x'}^t) |\theta_D).
    % \end{aligned}
    % \end{equation}
    
    % In this way,  HFC will be added to the input images. 
    % The diagram illustrating the generation of the adversarial samples and the influence on cross-domain gaze estimation is shown in \figref{fig:attack}.
    
% \input{fig/attack}
    
    % \subsection{Contrastively Reducing the Impact of HFC}
    \subsection{Contrastive Optimization}\label{sec:con}
    As described in \secref{sec:design}, we use contrastive learning to reduce the impact of HFC. 
    By encouraging the original and augmented samples to have similar representations, the network learns the ability to learn features with less impact of HFC.
    % By maintaining consistency between the original and adversarial samples, the network will learn the ability to reduce the impact of HFC.
    
    After the data augmentation, we have in total $4B$ samples $\{x_i^s, x_i^t, {x'}_i^s, {x'}_i^t |_{i=1}^B\}$. Taking a target sample $x_u^t$ for example, only $(x_u^t, {x'}_u^t)$ is treated as a positive pair for contrastive learning, the other $4B-2$ samples are considered negative ones.
    % This principle also applies to source domain samples. 
    Therefore, following the definition from \cite{C:chen2020simclr}, we define the contrastive loss $\ell_{con}(x_u, x_v)$ for a positive pair $(x_u, x_v)$ as:
    
    \begin{equation} \label{eq:conloss}
    \begin{aligned}
    &\ell_{con}(x_u, x_v; \theta_G) = \\
    &-\log \frac{\exp(sim(F(x_u| \theta_G), F(x_v| \theta_G))/\tau)}
    {\sum_{i=1}^{4B} \mathbbm{1}_{[i\neq u]}\exp(sim(F(x_u| \theta_G), F(x_i| \theta_G))/\tau)},
    \end{aligned}
    \end{equation}
    
    \noindent where $\tau$ is a temperature hyper-parameter \cite{C:wu2018unsupervised}, we empirically set $\tau=0.5$. The similarity measure $sim(\mathbf{f_u}, \mathbf{f_v})$ is defined with dot product as:
    
    \begin{equation} \label{eq:sim}
    \begin{aligned}
    sim(\mathbf{f_u}, \mathbf{f_v})=\frac{\mathbf{f_u} \bullet \mathbf{f_v}}
    {||\mathbf{f_u}|| \cdot ||\mathbf{f_v}||},
    \end{aligned}
    \end{equation}
    
    \noindent In \proposed, the total contrastive loss $\mathcal{L}_{con}(x^s, x^t, {x'}^s, {x'}^t)$ is computed over all positive pairs in a batch. 
    % In this way, the model learns the ability to reduce the impact of HFC via the optimization of the contrastive loss, resulting in better cross-domain performance.
    
    \subsection{Adversarial Domain Adaptation}\label{sec:adv}
    For better adaptation, we follow the idea of commonly-used adversarial learning to adapt the model to new domains. A domain discriminator $D(\cdot | \theta_D)$ is introduced, and the loss functions are defined as follows to encourage it to play a min-max game with the feature extractor $F$.
    
    \begin{equation} \label{eq:advloss}
    \begin{aligned}
    \underset {\theta_D}{\operatorname {arg\,min}}\,\mathcal{L}_{dis}(x^{s,t}, {x'}^{s,t}; \theta_D)= &-\log(1-D(F(x^s, {x'}^s) |\theta_D))\\
    &-\log D(F(x^t, {x'}^t) |\theta_D),\\
    \underset {\theta_G}{\operatorname {arg\,min}}\,\mathcal{L}_{adv}(x^t; \theta_G) = & -\log(1-D(F(x^t|\theta_G))).
    \end{aligned}
    \end{equation}
    
    The adversarial learning follows the classical procedure that are proved to be effective on domain adaptation tasks \cite{D:tzeng2017adda, D:cui2020gvbgd}. In summary, our goal of optimization on the gaze estimation network is defined as:
    
    \begin{equation} \label{eq:totalloss}
    \begin{aligned}
    \mathcal{L}= &\mathcal{L}_{gaze}(x^s, y^s; \theta_G) + \mathcal{L}_{gaze}({x'}^s, y^s; \theta_G)+\\
    &\lambda_1 \mathcal{L}_{con}(x^s, x^t, {x'}^s, {x'}^t; \theta_G)+\\
    &\lambda_2 (\mathcal{L}_{adv}(x^t; \theta_G)+\mathcal{L}_{adv}({x'}^t; \theta_G)),
    \end{aligned}
    \end{equation}
    where $\lambda_1$ and $\lambda_2$ are tunable parameters. We empirically set $\lambda_1$ = 1.0 and $\lambda_2$ = 0.1 in our experiments.

    \subsection{Adaptation Procedure}
    The adaptation procedure is summarized in Algorithm~\ref{alg:adapt}. 
    % The pre-trained network is randomly drawn from different epochs. 
    A small amount of data (\ie~ 100 images) with ground-truth labels from the source domain $\mathcal{D}_s$ and a small amount of data (\ie~ 100 images) from the target domain $\mathcal{D}_t$ are used for unsupervised adaptation. During adaptation, the network $G(\cdot|\theta_G)$ is trained by minimizing the loss function \eqnref{eq:totalloss}, and the domain discriminator $D(\cdot|\theta_D)$ is trained by minimizing the loss function \eqnref{eq:advloss}.

    \begin{algorithm}[t]
	\caption{Our gaze adaptation framework}
	\label{alg:adapt}
	
	\begin{algorithmic}[1]
		\Require {$G(\cdot|\theta_G^{(0)})$ pre-trained on source domain, small $\mathcal{D}_t$, and small $\mathcal{D}_s$}.
		\Ensure{$G(\cdot|\theta_G)$}
		
		\State Initialize: ${y}^t \gets G(x^t|\theta_G^{(0)})$; $D(\cdot|\theta_D^{(0)})$ \Comment{$\theta_D$ is randomly initialized.}

		\For{$ t{\gets}1\; to\; T $}
		
		\State $(x^s, y^s), x^t \gets \mathcal{D}_s, \mathcal{D}_t$
		
		\State ${x'}^s, {x'}^t \gets x^s, y^s, x^t, {y}^t, G, D $ with \eqnref{eq:img-aug}
		
		\State $\mathcal{L}_{gaze} \gets y^s, G(x^s|\theta_G^{(t-1)}), G({x'}^s|\theta_G^{(t-1)})$ with \eqnref{eq:gazeloss}
		
		\State $\mathcal{L}_{adv} \gets x^t, {x'}^t, G, D $ with \eqnref{eq:advloss}
		
		\State $\mathcal{L}_{con} \gets \frac{1}{4B}
		\sum_{i=1}^B[\ell_{con}(x_i^t, {x'}_i^t) + \ell_{con}({x'}_i^t, x_i^t) + \ell_{con}(x_i^s, {x'}_i^s) + \ell_{con}({x'}_i^s, x_i^s)]$ with \eqnref{eq:conloss}
		
		\State Update $G(\cdot|\theta_G^{(t)})$ with \eqnref{eq:totalloss}
		
		\State $\mathcal{L}_{dis} \gets x^s, x^t, {x'}^s, {x'}^t, G, D $ with \eqnref{eq:advloss}
		
		\State Update $D(\cdot|\theta_D^{(t)})$ with $\mathcal{L}_{dis}$
		
		\EndFor
		
% 		\State $\mu, \sigma \gets \mathcal{D}_v, G(\cdot|\theta_G^{(T)})$ with \eqnref{eq:deviation}
		
% 		\State $G(\cdot|\theta_G) \gets \mu, \sigma, G(\cdot|\theta_G^{(T)})$ with \eqnref{eq:vp}
	\end{algorithmic}
    \end{algorithm}

%% file: fig/overview.tex
\begin{figure*}[t]
\begin{center}
\includegraphics[width=\linewidth]{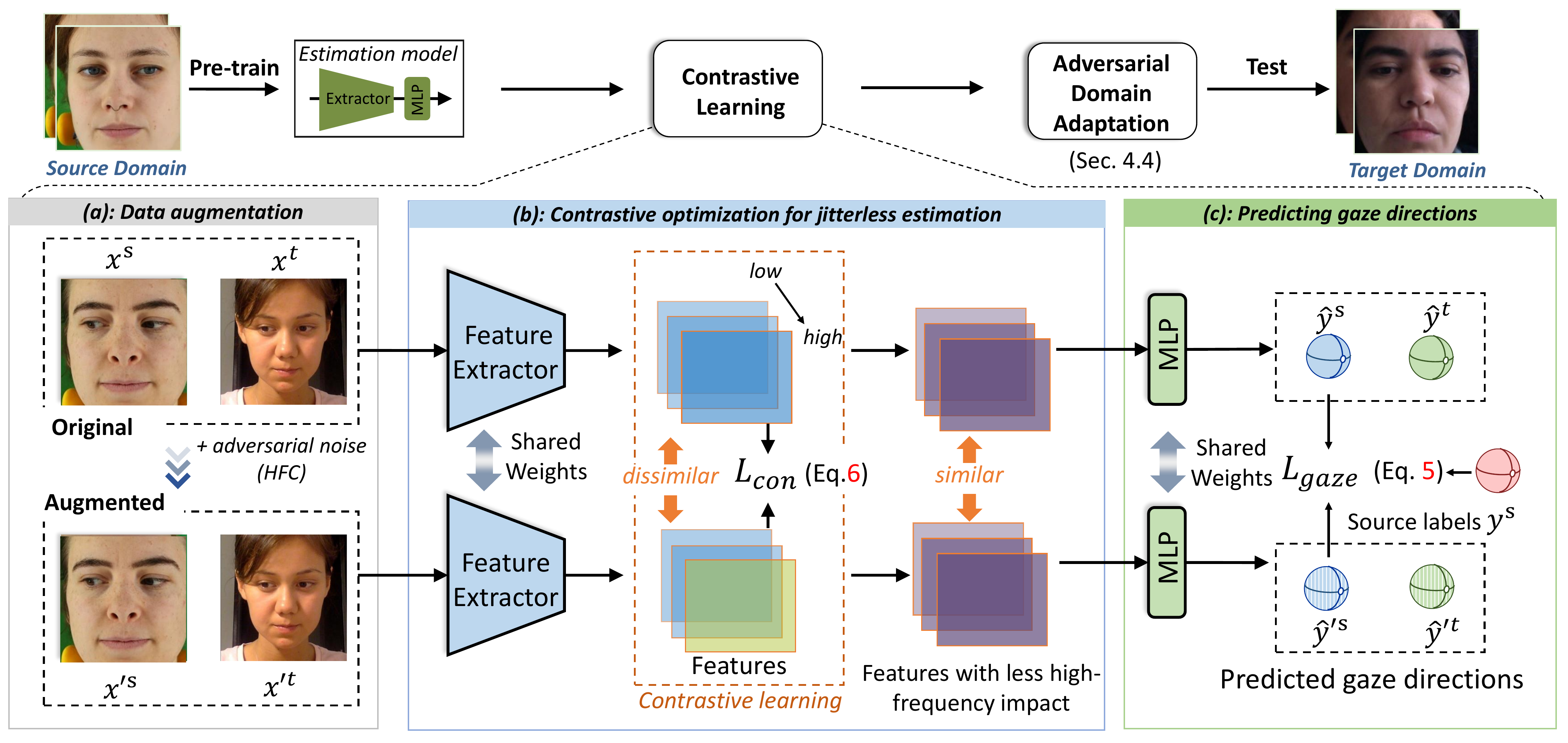}
\end{center}
\caption{
Overview of the proposed framework. (a) At first, pairs of augmented samples are generated by adding HFC with adversarial attack (\secref{sec:attack}). (b) Contrastive learning encourages the model to learn similar features from the original and adversarial samples, therefore reducing the impact of HFC (\secref{sec:con}). (c) Gaze directions are predicted, and the gaze labels from the source domain are employed for constructing the gaze loss.
}
\label{fig:overview}
% \vspace{20mm}
\end{figure*}

%% file: sec/5_experiment.tex
\section{Experiments}

\subsection{Data preparation}
Due to the variation in different datasets, data preparation is necessary. In our domain adaptation task, we utilize the ETH-XGaze (E) and Gaze360 (G) dataset as source domains, and MPIIGaze (M) and EyeDiap (D) dataset as target domains. 

The ETH-XGaze dataset~\cite{G:zhang2020eth}provides 80 subjects (\ie, 756,540 images), and we use them all as a source domain. For the Gaze360 dataset~\cite{G:kellnhofer2019gaze360}, we remove the images without subjects' faces and employ the remaining 112,251 images. For the MPIIGaze dataset\cite{G:zhang2017mpii}, we adopt the provided evaluation protocol to generate the evaluation set, which contains 3,000 images for each subject(\ie, 45,000 images).
% Since the MPIIGaze dataset\cite{G:zhang2017mpii} (MPII) provides a standard evaluation protocol, which selects 3000 images from each subject to form an evaluation set, we adopt the evaluation set directly(\ie, 45,000 images). 
For the EyeDiap dataset~\cite{G:funes2014eyediap}, we employ 16,674 images from 14 subjects under screen target sessions as the evaluation set.

Changes in head poses can significantly affect the appearance of face images. Therefore, we normalized the images following the method in~\cite{G:zhang2017mpii}, which ensures that the influence of head poses is eliminated.

% \subsection{Illustration of Jitter Level}
% Firstly, we use local-consistency which is previously defined to measure the level of jitter on both source and target domains. We directly apply models pre-trained on the source domains to the target domains, and the local-consistency results are shown in \tabref{tab:jitter}. The consistency sees a significant decrement when applying the models to the target domains, which also shows that the jitter is indeed more significant on the target domain.
% \input{tab/jitter}

% \subsection{Impact of High Frequency Information}
% \label{sec:low-fre}
% This experiment proves the assumption that HFC is one important factor of jitter on the target domain. In this section, the input images during testing are filtered out of HFC. We first adjust the proportion of information retained and observe the gaze accuracy, and the results are shown in \figref{fig:low-freq}. It can be observed that filtering out high-frequency will improve the gaze estimation performance, and the lowest point is approximately at 7\%. On this basis, we calculate the local-consistency at the lowest point and show the results in \tabref{tab:low-fre}. Both the local-consistency and gaze accuracy increase when using low-frequency images as input, which proves that HFC plays an important role in the jitter phenomena.
% \input{tab/low-fre}
% \input{fig/low-freq}

\subsection{Ablation Study} \label{sec:ablation-study}
% This section aims to show the optimality of our proposed framework.

\paragraph{Data Augmentation} In \secref{sec:attack}, we combine two adversarial attacks (FGSM and PGD) to add additive HFC to the input images. To prove the optimality, we conduct experiments to compare with other methods to add high-frequency noise. We employ random noises (Gaussian and Poisson) and adversarial noises (FGSM and PGD) to add high-frequency noise respectively. The results are shown in \tabref{tab:img-aug}. We can see that FGSM and PGD each achieve the best results on some tasks, so the final strategy is to use both as augmentation while adaptation (50\% probability of each being used), and the results are shown in the "Both" row.

Besides, in \secref{sec:high-freq}, directly removing high-frequency can also improve the performance. Therefore, it is also considered one possible data augmentation method in our experiment. So we directly use low-pass data instead of adding HFC during the data augmentation phase, and the results are shown in the "low-pass" row in \tabref{tab:img-aug}. The performance is not as good as adding additive HFC.

% Two augmentation directions are explored in this experiment: adding or removing HFC. Specifically, when performing the adding way, we test both random noises (Gaussian and Poisson) and adversarial noises (FGSM and PGD). The results are shown in \tabref{tab:img-aug}. We find that FGSM and PGD each achieve the best results on some tasks, so we simultaneously used both as augmentation while adaptation (50\% probability of each being used), and the results are shown in the "2-way" row. The combination of FGSM and PGD reaches the best performance among all four adaptation tasks.
\input{tab/img-aug}

\paragraph{Contrastive Learning} In this section, we conduct experiments to explore how contrastive loss affects \proposed's performance.
Recall that in \secref{sec:con}, we adopt the loss setting from SimCLR~\cite{C:chen2020simclr} for implementation.
To find the suitable contrastive loss, we further replace this part with the loss settings from the other two commonly-used frameworks MoCo~\cite{C:he2020moco} and PCL~\cite{C:li2020pcl}, respectively.

To this end, we rebuild \proposed~ with different contrastive modules, then adapt these modified versions under the same condition.
% In this section, we keep using the same adversarial attack to generate augmented samples, but build different contrastive modules for keeping the consistency between raw and adversarial samples. 
Specifically, for the contrastive loss from MoCo, we utilize a temporal average model to generate the contrastive loss, then we term this version as Ours-MoCo. 
When it comes to PCL, we follow the settings from its original paper and introduce clustering to generate the contrastive loss. This version is named Ours-PCL.
For a fair comparison, each of these models uses 100 target samples for UDA, and tests on the whole target domain.

One important indicator of the contrastive learning module is the ability to reduce the impact of HFC. 
To compare this ability, we use triplet loss\cite{N:schroff2015facenet} as an evaluation metric to measure the feature difference between the original and adversarial data.
\figref{fig:triplet} shows the triplet loss, which reflects the capability of different modules for keeping the consistency before and after being added HFC. 
In this figure,  we find \proposed~ gets the consistently lowest triplet loss, even with the variation of margins, which means it better reduces the impact of HFC. 
% In other words, it shows that our module always produces the best consistency results, even with the variation of margins. 

Furthermore, we compare the cross-domain gaze estimation performances with these different contrastive learning modules. Quantitative results are shown in \tabref{tab:contrastive}. Statistical results consistently show that \proposed~ also achieves the highest stability and accuracy on these four tasks.

\input{fig/triplet}
\input{tab/contrastive}

\paragraph{Ablation} %We test the effect of each module in our proposed method \proposed.
An ablation study is conducted to demonstrate the effectiveness of each component from \proposed. The components are shown below.
\begin{itemize}[leftmargin=*]
\setlength\itemsep{-.3em}
\item CNN: A CNN gaze estimation network using ResNet18\cite{N:he2016deep}. The network is pre-trained on the source domain.
\item adv: Adversarial domain adaptation module, which is used to minimize the distance between the source and target domains.
\item con: Contrastive learning module, which reduces the impact of HFC by keeping the consistency between original and adversarial samples.
\end{itemize}
For all the experiments, we load the model trained on the source domain as initial weights. During adaptation, 100 source samples and 100 target samples are used. \tabref{tab:ablation} shows the \metric~ and accuracy under different combinations. 

It is clear that common adversarial domain adaptation (CNN+adv) has a limited cross-domain performance improvement. 
After adding the proposed contrastive learning module,  our method achieves the best results on all these 4 tasks, which confirms the effectiveness of reducing the impact of HFC.

\input{tab/ablation}

\subsection{Resist to High-Frequency Noise}

As announced in \secref{sec:con}, the contrastive learning module reduces the impact of HFC. To further verify the announcement, we conduct experiments to add high-frequency noise to the test data and compare the impact on the baseline and adapted model brought by the noise. 

Specifically, we utilize Gaussian/Poisson noise as additive high-frequency noise and add them to the target data. We take the pre-trained model and adapted models and compare their \metric~ and gaze accuracy on the "perturbed" target domain. Quantitative results are shown in \tabref{tab:anti-jitter}. We find that our method outperforms the baseline, both in comparing their values and their changes in value, which means \proposed~ does help the model reduce the impact of HFC.

\input{tab/anti-jitter}

\subsection{Performance in Source Domains}

In this section, we conduct experiments to verify the performance of the model on the source domain after adaptation to the target domain.

We directly test the model, which has been adpated to the target domain, in the source domain, and the experimental results are shown in \tabref{tab:toself}. It can be seen that the \metric~ and gaze error of the adapted model increase only slightly, and the \metric~ even decreases on the Gaze360 dataset. This indicates that our method can also continue to maintain its performance in source domains after adaptation.

\input{tab/toself}

\subsection{Comparison with SOTA UDA Methods}

We also compare the UDA performances between \proposed~ and other state-of-the-art UDA methods.
% \cite{D:guo2020dagen, D:tzeng2017adda, D:wang2019gazeadv, G:kellnhofer2019gaze360, D:cui2020gvbgd}.

DAGEN \cite{D:guo2020dagen}, GazeAdv \cite{D:wang2019gazeadv}, Gaze360 \cite{G:kellnhofer2019gaze360}, PnP-GA \cite{D:liu2021generalizing}, and PureGaze \cite{G:cheng2022puregaze} propose different methods for cross-domain gaze estimation.
It is worth noting that PnP-GA requires a specific group of models (10 models) for effective domain adaptation, so the method is under different settings from others.
In contrast, the ADDA\cite{D:tzeng2017adda} and GVBGD \cite{D:cui2020gvbgd} were originally proposed for classification tasks. Here, we conduct experiments with these two methods to show the performance of state-of-the-art UDA methods on cross-domain gaze estimation. All the backbones are replaced with ResNet18, which is the same as ours for a fair comparison. In detail, we also adjust the number of samples to reach the best performance of these methods.

Quantitative results are shown in \tabref{tab:sota}. The results of fine-tuning are shown as the upper bound of adaptation, it uses 100 samples which is the same as ours. We found that our method significantly outperforms the state-of-the-art UDA methods. The superior performance of our method validates the effectiveness of \proposed~ for cross-domain gaze estimation.

\input{tab/sota}

%% file: tab/img-aug.tex
\begin{table} 
\caption{Performance comparison of \proposed~ using different image augmentation methods.}
\centering
\resizebox{\linewidth}{!}{
\begin{tabular}{lcccccccc}  
\toprule[1.5pt]
		
 & \multicolumn{2}{c}{\underline{E$ \rightarrow $M}} & \multicolumn{2}{c}{\underline{E$\rightarrow $D}} &
\multicolumn{2}{c}{\underline{G$\rightarrow $M}} & \multicolumn{2}{c}{\underline{G$\rightarrow $D}} \\ 
		&\metric   & $error$  & \metric & $error$ & \metric & $error$ & \metric & $error$  \\ 
\midrule[1pt] 
		
Baseline	 	& 3.99 & 8.56 & 6.65 & 8.60 & 4.38 & 9.52 & 6.00 & 10.05 \\  
\midrule
Low-pass	& 3.67 & 7.19 & 5.85 & 7.97 & 5.82 & 12.33 & 5.49 & 17.41 \\
\midrule
G$\sim$0.01	& 3.07 & 6.01 & 5.23 & 6.63 & 4.37 & 8.42 & 5.44 & 9.66 \\
G$\sim$0.05	& 3.19 & 6.46 & 5.69 & 6.83 & 4.56 & 8.65 & 5.60 & 9.17 \\
P$\sim$10	& 3.15 & 6.48 & 4.57 & 6.65 & 4.57 & 8.72 & 5.62 & 9.28 \\
P$\sim$15	& 3.05 & 6.25 & 6.12 & 7.51 & 4.56 & 9.00 & 5.83 & 9.57 \\
\midrule
FGSM    & \textbf{2.09} & 5.41 & 5.01 & \textbf{6.62} & 3.66 & 7.28 & \textbf{4.41} & 8.62 \\
PGD    & 2.19 & \textbf{5.31} & 5.17 & 6.83 & 3.54 & 7.24 & 4.58 & 8.94 \\
Both    & 2.21 & 5.35 & \textbf{4.52} & \textbf{6.62} & \textbf{3.51} & \textbf{7.18} & \textbf{4.41} & \textbf{8.61} \\
		
		\bottomrule[1.5pt]
	\end{tabular}
}   
\label{tab:img-aug}
\end{table}

% \begin{table} 
% \caption{Performance comparison of our \proposed~ using different image augmentation methods.}
% \centering
% \resizebox{\linewidth}{!}{
% \begin{tabular}{lcccccccc}  
% \toprule[1.5pt]
		
%  & \multicolumn{2}{c}{\underline{E$ \rightarrow $M}} & \multicolumn{2}{c}{\underline{E$\rightarrow $D}} &
% \multicolumn{2}{c}{\underline{G$\rightarrow $M}} & \multicolumn{2}{c}{\underline{G$\rightarrow $D}} \\ 
% 		&\metric   & $error$  & \metric & $error$ & \metric & $error$ & \metric & $error$  \\ 
% \midrule[1pt] 
		
% Baseline	 	& 3.75 & 8.56 & 6.65 & 8.60 & 4.38 & 9.52 & 6.00 & 10.05 \\  
% \midrule
% Low-freq.	& 3.56 & 7.19 & 5.85 & 7.97 & 5.82 & 12.33 & 5.49 & 17.41 \\
% \midrule
% G$\sim$0.01	& 3.18 & 6.01 & 5.23 & 6.63 & 4.37 & 8.42 & 5.44 & 9.66 \\
% G$\sim$0.05	& 3.23 & 6.46 & 5.69 & 6.83 & 4.56 & 8.65 & 5.60 & 9.17 \\
% P$\sim$10	& 3.26 & 6.48 & 4.57 & 6.65 & 4.57 & 8.72 & 5.62 & 9.28 \\
% P$\sim$15	& 3.19 & 6.25 & 6.12 & 7.51 & 4.56 & 9.00 & 5.83 & 9.57 \\
% \midrule
% FGSM    & \textbf{2.22} & 5.41 & 5.01 & \textbf{6.62} & 3.66 & 7.28 & \textbf{4.41} & 8.62 \\
% PGD    & 2.27 & \textbf{5.31} & 5.17 & 6.83 & 3.54 & 7.24 & 4.58 & 8.94 \\
% 2-Way    & \textbf{2.22} & 5.35 & \textbf{4.52} & \textbf{6.62} & \textbf{3.51} & \textbf{7.18} & \textbf{4.41} & \textbf{8.61} \\
		
% 		\bottomrule[1.5pt]
% 	\end{tabular}
% }   
% \label{tab:low-fre}
% \end{table}

%% file: fig/triplet.tex
\begin{figure}[t]
\begin{center}
\includegraphics[width=\linewidth]{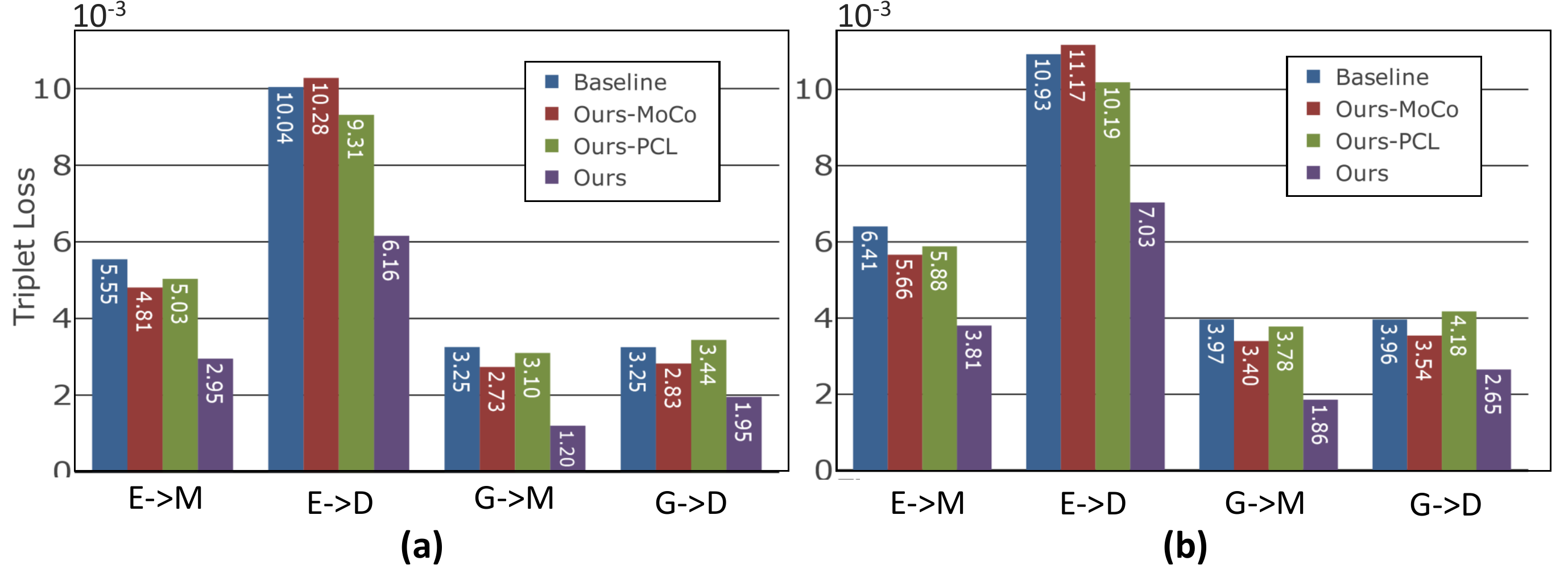}
\end{center}
\caption{
Comparison with other contrastive learning modules on triplet loss with different margins. Lower loss means better reducing the impact of HFC. The margin is set to 0 and $1\times10^{-3}$ in (a) and (b), respectively. Our method is consistently the best.
}
\label{fig:triplet}
\end{figure}

%% file: tab/contrastive.tex
\begin{table} 
% \vspace{10mm}
\caption{Performance comparison of \proposed~ using different contrastive learning modules.}
\centering
\resizebox{\linewidth}{!}{
\begin{tabular}{lcccccccc}  
\toprule[1.5pt]
		
 & \multicolumn{2}{c}{\underline{E$ \rightarrow $M}} & \multicolumn{2}{c}{\underline{E$\rightarrow $D}} &
\multicolumn{2}{c}{\underline{G$\rightarrow $M}} & \multicolumn{2}{c}{\underline{G$\rightarrow $D}} \\ 
		&\metric   & $error$  & \metric & $error$ & \metric & $error$ & \metric & $error$  \\ 
\midrule[1pt] 
		
Baseline	 	& 3.99 & 8.56 & 6.65 & 8.60 & 4.38 & 9.52 & 6.00 & 10.05 \\  
\midrule
Ours-MoCo    & 3.17 & 6.48 & 7.06 & 9.42 & 4.79 & 8.74 & 5.59 & 11.60 \\
Ours-PCL    & 3.25 & 6.38 & 6.66 & 8.52 & 4.66 & 8.85 & 5.35 & 10.31 \\
Ours    & \textbf{2.21} & \textbf{5.35} & \textbf{4.52} & \textbf{6.62} & \textbf{3.51} & \textbf{7.18} & \textbf{4.41} & \textbf{8.61} \\
		
	\bottomrule[1.5pt]
	\end{tabular}
}   
\label{tab:contrastive}
\end{table}

% \begin{table} 
% \caption{Performance comparison of our Ours~ using different contrastive learning methods.}
% \centering
% \resizebox{\linewidth}{!}{
% \begin{tabular}{lcccccccc}  
% \toprule[1.5pt]
		
%  & \multicolumn{2}{c}{\underline{E$ \rightarrow $M}} & \multicolumn{2}{c}{\underline{E$\rightarrow $D}} &
% \multicolumn{2}{c}{\underline{G$\rightarrow $M}} & \multicolumn{2}{c}{\underline{G$\rightarrow $D}} \\ 
% 		&\metric   & $error$  & \metric & $error$ & \metric & $error$ & \metric & $error$  \\ 
% \midrule[1pt] 
		
% Baseline	 	& 3.75 & 8.56 & 6.65 & 8.60 & 4.38 & 9.52 & 6.00 & 10.05 \\  
% \midrule
% Ours-MoCo    & 3.19 & 6.48 & 7.06 & 9.42 & 4.79 & 8.74 & 5.59 & 11.60 \\
% Ours-PCL    & 3.23 & 6.38 & 6.66 & 8.52 & 4.66 & 8.85 & 5.35 & 10.31 \\
% Ours    & \textbf{2.22} & \textbf{5.35} & \textbf{4.52} & \textbf{6.62} & \textbf{3.51} & \textbf{7.18} & \textbf{4.41} & \textbf{8.61} \\
		
% 	\bottomrule[1.5pt]
% 	\end{tabular}
% }   
% \label{tab:low-fre}
% \end{table}

%% file: tab/ablation.tex
\begin{table} 
\caption{Ablation study of different components in \proposed.}
\centering
\resizebox{\linewidth}{!}{
\begin{tabular}{lcccccccc}  
\toprule[1.5pt]
		
 & \multicolumn{2}{c}{\underline{E$ \rightarrow $M}} & \multicolumn{2}{c}{\underline{E$\rightarrow $D}} &
\multicolumn{2}{c}{\underline{G$\rightarrow $M}} & \multicolumn{2}{c}{\underline{G$\rightarrow $D}} \\ 
		&\metric   & $error$  & \metric & $error$ & \metric & $error$ & \metric & $error$  \\ 
\midrule[1pt] 
		
Baseline	 	& 3.99 & 8.56 & 6.65 & 8.60 & 4.38 & 9.52 & 6.00 & 10.05 \\ 
\midrule
CNN+con    & 2.34 & 5.37 & 5.16 & 6.58 & 3.79 & \textbf{7.18} & 4.52 & 8.62 \\
CNN+adv    & 3.41 & 7.24 & 6.43 & 7.59 & 4.11 & 7.72 & 5.22 & 9.86 \\
CNN+con+adv    & \textbf{2.21} & \textbf{5.35} & \textbf{4.52} & \textbf{6.62} & \textbf{3.51} & \textbf{7.18} & \textbf{4.41} & \textbf{8.61} \\
		
	\bottomrule[1.5pt]
	\end{tabular}
}   
\label{tab:ablation}
\end{table}

% \begin{table} 
% \caption{Ablation study of different components in our proposed \proposed.}
% \centering
% \resizebox{\linewidth}{!}{
% \begin{tabular}{lcccccccc}  
% \toprule[1.5pt]
		
%  & \multicolumn{2}{c}{\underline{E$ \rightarrow $M}} & \multicolumn{2}{c}{\underline{E$\rightarrow $D}} &
% \multicolumn{2}{c}{\underline{G$\rightarrow $M}} & \multicolumn{2}{c}{\underline{G$\rightarrow $D}} \\ 
% 		&\metric   & $error$  & \metric & $error$ & \metric & $error$ & \metric & $error$  \\ 
% \midrule[1pt] 
		
% Baseline	 	& 3.75 & 8.56 & 6.65 & 8.60 & 4.38 & 9.52 & 6.00 & 10.05 \\ 
% \midrule
% CNN+con    & 2.42 & 5.37 & 5.16 & 6.58 & 3.79 & \textbf{7.18} & 4.52 & 8.62 \\
% CNN+adv    & 3.36 & 7.24 & 6.43 & 7.59 & 4.11 & 7.72 & 5.22 & 9.86 \\
% CNN+con+adv    & \textbf{2.22} & \textbf{5.35} & \textbf{4.52} & \textbf{6.62} & \textbf{3.51} & \textbf{7.18} & \textbf{4.41} & \textbf{8.61} \\
		
% 	\bottomrule[1.5pt]
% 	\end{tabular}
% }   
% \label{tab:low-fre}
% \end{table}

%% file: tab/anti-jitter.tex
\begin{table} 
\caption{Comparison of the impact of high-frequency information on the baseline and adapted model. Our method outperforms the baseline, both in values and changes in value. }
% \vspace{10mm}
\centering
\resizebox{0.85\linewidth}{!}{
\begin{tabular}{lccccc}  
\toprule[1.5pt]
		
\multirow{2}{1cm}{Noise} & \multirow{2}{1cm}{Model} & \multicolumn{2}{c}{\underline{E$ \rightarrow $M}} &
\multicolumn{2}{c}{\underline{G$\rightarrow $M}}  \\ 
		& &\metric   & $error$  & \metric & $error$  \\ 
\midrule[1pt] 
		
\multirow{2}{1cm}{G$\sim$0.01} & Baseline & 4.10 & 8.79 & 4.47 & 9.60 \\ 
    & Ours & \textbf{2.25} & \textbf{5.36} & \textbf{3.53} & \textbf{7.14} \\

\midrule[1pt] 

\multirow{2}{1cm}{G$\sim$0.05} & Baseline & 6.44 & 10.61 & 5.96 & 10.77 \\ 
    & Ours & \textbf{4.21} & \textbf{6.37} & \textbf{4.48} & \textbf{6.99} \\

\midrule[1pt] 

\multirow{2}{1cm}{P$\sim$10} & Baseline & 5.76 & 10.12 & 5.58 & 10.50 \\ 
    & Ours & \textbf{3.56} & \textbf{6.09} & \textbf{4.26} & \textbf{6.99} \\

\midrule[1pt] 

\multirow{2}{1cm}{P$\sim$15} & Baseline & 7.24 & 11.31 & 6.47 & 11.26 \\ 
    & Ours & \textbf{4.73} & \textbf{6.92} & \textbf{5.04} & \textbf{7.06} \\

	\bottomrule[1.5pt]
	\end{tabular}
}   
\label{tab:anti-jitter}
\end{table}

% \begin{table} 
% \caption{Performace comparison under different perturbations. The percentage value indicates the increase/decrease compared to the gaze error before ”perturbed”.}
% \centering
% \resizebox{\linewidth}{!}{
% \begin{tabular}{lccccc}  
% \toprule[1.5pt]
		
%  & & \multicolumn{2}{c}{\underline{E$ \rightarrow $M}} &
% \multicolumn{2}{c}{\underline{G$\rightarrow $M}}  \\ 
% 		& &\metric   & $error$  & \metric & $error$  \\ 
% \midrule[1pt] 
		
% \multirow{2}{1cm}{G$\sim$0.01} & Baseline & 3.80 & 8.66 & 3.79 & 8.66 \\ 
%     & Ours & 2.28 & 5.36 & 3.53 & 7.14 \\

% \midrule

% \multirow{2}{1cm}{G$\sim$0.05} & Baseline & 6.23 & 10.59 & 6.17 & 10.59 \\ 
%     & Ours & 4.70 & 6.36 & 4.48 & 6.99 \\

% \midrule

% \multirow{2}{1cm}{P$\sim$10} & Baseline & 5.49 & 10.12 & 5.55 & 10.12 \\ 
%     & Ours & 4.07 & 6.09 & 4.26 & 6.99 \\

% \midrule

% \multirow{2}{1cm}{P$\sim$15} & Baseline & 6.73 & 11.29 & 6.78 & 11.31 \\ 
%     & Ours & 5.78 & 6.92 & 5.04 & 7.06 \\

% 	\bottomrule[1.5pt]
% 	\end{tabular}
% }   
% \label{tab:low-fre}
% \end{table}

%% file: tab/toself.tex
\begin{table} 
\caption{Performance of \proposed~ in the source domains after adaptation to the target domains.}
\centering
%\resizebox{\linewidth}{!}{
\begin{tabular}{lcccc}  
\toprule[1.5pt]
		
 & \multicolumn{2}{c}{\underline{ETH-XGaze}} & \multicolumn{2}{c}{\underline{Gaze360}} \\ 
		&\metric   & $error$  & \metric & $error$  \\ 
\midrule[1pt] 
		
Baseline	 	& \textbf{0.71} & \textbf{4.42} & 3.08 & \textbf{11.59} \\ 
\midrule
$\rightarrow$MPIIGaze    & 0.75 & 5.19 & \textbf{2.21} & 11.86 \\
$\rightarrow$EyeDiap    & 0.77 & 5.18 & 2.51 & 11.98 \\
		
	\bottomrule[1.5pt]
	\end{tabular}
%}   
\label{tab:toself}
\end{table}

% \begin{table} 
% \caption{Ablation study of different components in our proposed \proposed.}
% \centering
% \resizebox{\linewidth}{!}{
% \begin{tabular}{lcccccccc}  
% \toprule[1.5pt]
		
%  & \multicolumn{2}{c}{\underline{E$ \rightarrow $M}} & \multicolumn{2}{c}{\underline{E$\rightarrow $D}} &
% \multicolumn{2}{c}{\underline{G$\rightarrow $M}} & \multicolumn{2}{c}{\underline{G$\rightarrow $D}} \\ 
% 		&\metric   & $error$  & \metric & $error$ & \metric & $error$ & \metric & $error$  \\ 
% \midrule[1pt] 
		
% Baseline	 	& 3.75 & 8.56 & 6.65 & 8.60 & 4.38 & 9.52 & 6.00 & 10.05 \\ 
% \midrule
% CNN+con    & 2.42 & 5.37 & 5.16 & 6.58 & 3.79 & \textbf{7.18} & 4.52 & 8.62 \\
% CNN+adv    & 3.36 & 7.24 & 6.43 & 7.59 & 4.11 & 7.72 & 5.22 & 9.86 \\
% CNN+con+adv    & \textbf{2.22} & \textbf{5.35} & \textbf{4.52} & \textbf{6.62} & \textbf{3.51} & \textbf{7.18} & \textbf{4.41} & \textbf{8.61} \\
		
% 	\bottomrule[1.5pt]
% 	\end{tabular}
% }   
% \label{tab:low-fre}
% \end{table}

%% file: tab/sota.tex
\begin{table}
% \scriptsize
\caption{
Comparison with state-of-the-art unsupervised domain adaptation approaches. $^\dag$ indicates that target gaze labels are used. $^\ddag$ indicates that experimental settings are different. $^*$  indicates that more than 100 target samples are used during adaptation. 
% Angular gaze error($^\circ$) is used as evaluation metric.
% 
}
% \caption
\centering
\resizebox{\linewidth}{!}{ %< auto-adjusts font size to fill line
\begin{tabular}{lcccccccc}  
\toprule[1.5pt]
		
 & \multicolumn{2}{c}{\underline{E$ \rightarrow $M}} & \multicolumn{2}{c}{\underline{E$\rightarrow $D}} &
\multicolumn{2}{c}{\underline{G$\rightarrow $M}} & \multicolumn{2}{c}{\underline{G$\rightarrow $D}} \\ 
		&\metric   & $error$  & \metric & $error$ & \metric & $error$ & \metric & $error$  \\ 
\midrule[1pt]

Baseline	 	& 3.99 & 8.56 & 6.65 & 8.60 & 4.38 & 9.52 & 6.00 & 10.05 \\ 
Fine-tune$^\dag$ & 2.73 & 4.37 & 5.35 & 5.64 & 3.20 & 5.63 & 4.38 & 5.74 \\
\midrule[1pt] 
PnP-GA$^\ddag$ & 2.99 & 5.53 & 3.83 & 5.87 & 2.76 & 6.18 & 5.31 & 7.92 \\
\midrule[1pt] 
ADDA$^*$ & 2.96 & 7.54 & 5.11 & 7.02 & 3.87 & 8.16 & 6.10 & 11.38 \\
DAGEN$^*$ & 2.91 & 5.87 & 6.56 & 9.81 & 4.23 & 7.70 & 4.85 & 12.18 \\
Gaze360 & 3.28 & 5.87 & 6.31 & 7.38 & 3.98 & 7.42 & 4.89 & 9.28 \\
GazeAdv & 3.66 & 7.61 & 6.50 & 8.27 & 4.46 & 8.21 & 5.63 & 10.68 \\
GVBGD$^*$ & 3.00 & 6.68 & 5.66 & 7.27 & 4.67 & 8.39 & 5.23 & 12.44 \\
PureGaze & 3.88 & 7.08 & 5.97 & 7.48 & 6.16 & 9.28 & 5.50 & 9.32 \\
Ours & \textbf{2.21} & \textbf{5.35} & \textbf{4.52} & \textbf{6.62} & \textbf{3.51} & \textbf{7.18} & \textbf{4.41} & \textbf{8.61} \\
\bottomrule[1.2pt]
\end{tabular}
} % \resizebox

\label{tab:sota}
\end{table}

%% file: sec/6_conclusion.tex
\section{Conclusion}

In this paper, we present a novel framework for adapting gaze estimation to new domains. We start by analyzing the gaze jitter phenomenon that occurs when crossing domains, and discover that HFC is one important factor leading to gaze jitter. This factor guides us to design the proposed method. Extensive experiments demonstrate the superior performance of \proposed~ for cross-domain gaze estimation tasks.
Our method has the potential to be used in real-world gaze estimation applications.